%% file: main.tex
\documentclass[runningheads]{llncs}

\usepackage{eccv}

\usepackage{eccvabbrv}

\usepackage{graphicx}
\usepackage{booktabs}

\usepackage[accsupp]{axessibility}  

\usepackage[pagebackref,breaklinks,colorlinks,citecolor=eccvblue]{hyperref}

\usepackage{orcidlink}

\usepackage{booktabs}       
\usepackage{amsfonts}       
\usepackage{nicefrac}       

\usepackage{xcolor}         

\usepackage{bm}
\usepackage{amsmath}

\usepackage{graphicx}
\usepackage{float}
\usepackage{wrapfig}

\usepackage{enumitem} 
\usepackage{algorithm}
\usepackage{algorithmic}

\usepackage{setspace}
\usepackage{multicol}
\usepackage{multirow}
\usepackage{colortbl}
\usepackage{makecell}
\usepackage{wrapfig} 
\usepackage{pifont} 
\usepackage{amssymb}

\definecolor{aliceblue}{rgb}{0.94, 0.97, 1.0}

\usepackage{comment}
\usepackage[dvipsnames]{xcolor}

\begin{document}

\title{ParCo: Part-Coordinating Text-to-Motion Synthesis} 


\author{Qiran Zou\thanks{Equal contribution.}\inst{1} \and
Shangyuan Yuan$^\star$\inst{1}\orcidlink{0009-0006-8832-6372} \and
Shian Du\inst{1} \and
Yu Wang\inst{2} \and
Chang Liu\inst{1} \and
Yi Xu\inst{3} \and
Jie Chen\inst{2} \and
Xiangyang Ji\inst{1}}

\authorrunning{Q.~Zou et al.}

\institute{Tsinghua University, China \\
\and
Peking University Shenzhen Graduate School, China \and 
Dalian Unversity of Technology, China\\
\email{qiranzou@gmail.com},
\email{\{yuansy21, dsa23\}@mails.tsinghua.edu.cn},
\email{2201212856@stu.pku.edu.cn},
\email{liuchang2022@tsinghua.edu.cn},
\email{yxu@dlut.edu.cn},
\email{chenj@pcl.ac.cn},
\email{xyji@tsinghua.edu.cn}
}

\maketitle
\begin{abstract}
We study a challenging task: text-to-motion synthesis, aiming to generate motions that align with textual descriptions and exhibit coordinated movements. Currently, the part-based methods introduce part partition into the motion synthesis process to achieve finer-grained generation. However, these methods encounter challenges such as the lack of coordination between different part motions and difficulties for networks to understand part concepts. Moreover, introducing finer-grained part concepts poses computational complexity challenges.
In this paper, we propose Part-Coordinating Text-to-Motion Synthesis (ParCo), endowed with enhanced capabilities for understanding part motions and communication among different part motion generators, ensuring a coordinated and fined-grained motion synthesis. 
Specifically, we discretize whole-body motion into multiple part motions to establish the prior concept of different parts. 
Afterward, we employ multiple lightweight generators designed to synthesize different part motions and coordinate them through our part coordination module. 
Our approach demonstrates superior performance on common benchmarks with economic computations, including HumanML3D and KIT-ML, providing substantial evidence of its effectiveness. 
Code is available at: \url{https://github.com/qrzou/ParCo}.
\end{abstract}


\section{Introduction}
\label{sec:intro}
Text-to-motion synthesis aims to generate motion that aligns with textual descriptions and exhibits coordinated movements.
It facilitates obtaining desired motion through textual descriptions which benefits numerous applications in industrial scenarios such as animation~\cite{kappel2021high}, AR/VR applications, video games~\cite{majoe2009enhanced, yeasin2004multiobject}, autonomous driving~\cite{djuric2020uncertainty}, and robotics~\cite{koppula2013learning, koppula2015anticipating, antakli2018intelligent}. 

\begin{center}
    \centering
    \captionsetup{type=figure}
        \includegraphics[width=\linewidth]{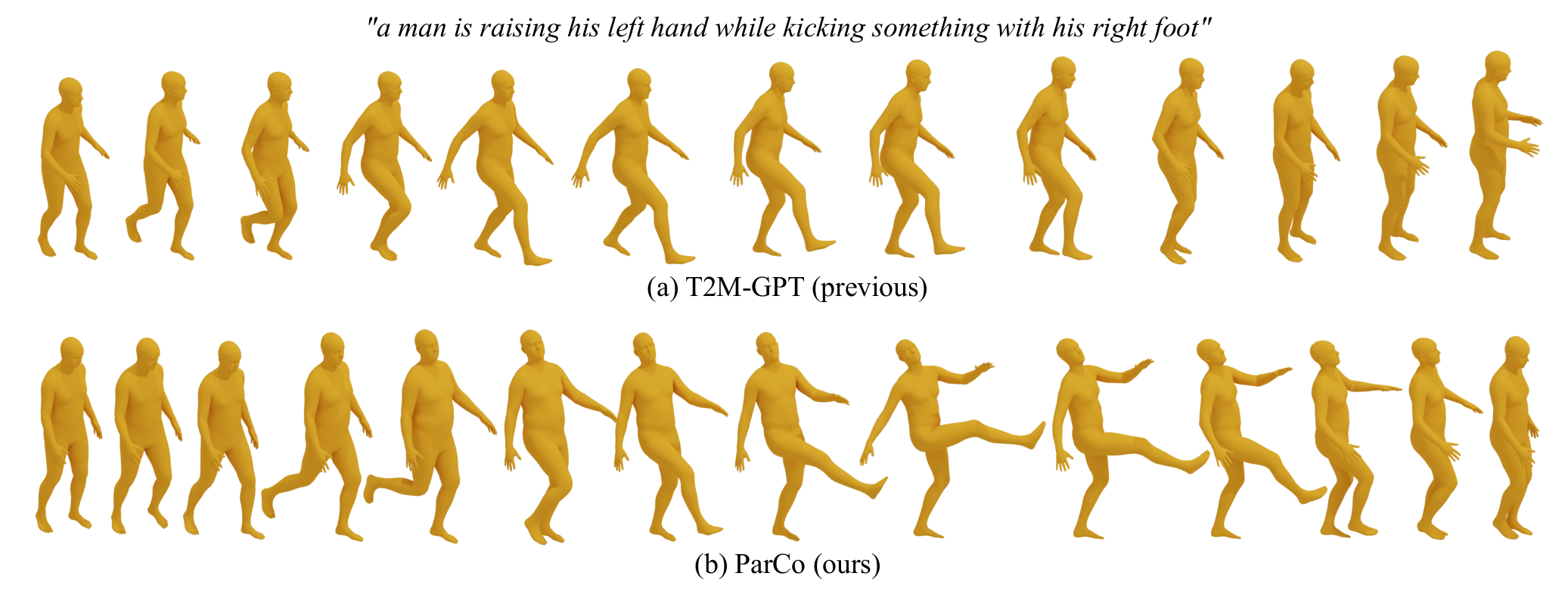}
        \captionof{figure}{
        Our ParCo is capable of coordinating the motion of various body parts to produce realistic and accurate motion.
        }
    \label{fig:intro_compare_results}
\end{center}

Recent advancements leveraging powerful generation capabilities of transformers~\cite{vaswani2017attention,parmar2018image,liu2021swin} and diffusion models~\cite{nichol2021glide,ramesh2022hierarchical,saharia2022photorealistic,rombach2022high,ho2020denoising} have yielded impressive results in generating realistic and smooth motions. 
Despite this progress, existing methods often struggle to generate semantically matched and coherently coordinated motions, especially when faced with commands involving multiple coordinated body parts. 
We attribute this challenge to the intricate alignment problem between the modalities of text and motion, where a single text/motion can correspond to multiple possible motion/text, posing a challenge in learning the complex relationship between the two.

\begin{figure}[!t]
   \centering
   \includegraphics[width=\linewidth]{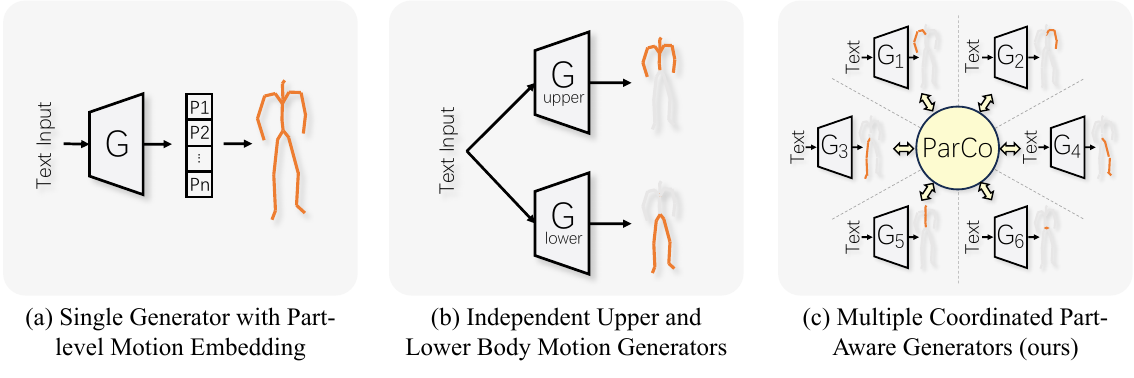}
   \caption{
        Conceptual comparison of three part-based synthesis methods. 
        \textbf{(a):} One generator synthesizes the whole-body embedding, which contains information about different parts internally.
        \textbf{(b):} Two separate generators synthesize the upper and lower body's motions independently, without information exchange between them. 
        \textbf{(c):} Our ParCo employs multiple lightweight generators designed to synthesize different part motions, which are coordinated by the Part Coordination module.
        }
   \label{fig:intro_compare_methods}
\end{figure}

In the realm of Text-to-Motion Generation, part-based methods aim to achieve a higher level of sophistication in motion generation. These approaches can be broadly classified into two categories: single generator with part-level motion embeddings (Fig.~\ref{fig:intro_compare_methods} (a)) and independent upper and lower body motion generators (Fig.~\ref{fig:intro_compare_methods} (b)).
The former one constructs whole-body motion embeddings by amalgamating multiple part motion embeddings, explicitly introducing the concept of parts~\cite{siyao2022bailando}, or implicitly integrating part concepts with part-level attention~\cite{zhong2023attt2m}. 
However, utilizing a single generator to generate whole-body motion embeddings presents a challenge for the generator to understand the concept of parts.
The latter one segregates whole-body motion into upper and lower body motions, deploying two independent generators to produce motions separately~\cite{ghosh2021synthesis}. Although this design enhances the generator's comprehension of upper and lower body motions, the absence of information exchange between them leads to a lack of coordination in the resulting upper and lower body motions. To increase the granularity of part division, an intuitive solution is to employ additional generators for generating finer-grained part motions, such as the left leg. However, this approach introduces computational complexity challenges.

Neuroscientific discoveries reveal that discrete regions within the human brain manifest unique functions, and these regions engage in communication to coordinate diverse activities~\cite{thiebaut2022emergent,herbet2020revisiting}.
This design, where different low-level subsystems communicate to form a higher-level system, is prevalent in the natural and artificial world, offering a robust, perceptually strong structural design.
In adherence to these principles, we present Part-Coordinating Text-to-Motion Synthesis (\textbf{ParCo}). It comprises six small generators, which are tasked with various part motions, and accompanied by a Part Coordination module facilitating communication among the generators. This communication enables the generation of coordinated whole-body motions while comprehending distinct parts (Fig.~\ref{fig:intro_compare_methods} (c)).
Thanks to the design of small generators, our method demonstrates a lower parameter count, reduced computational complexity, and shorter generation time in comparison to baseline and state-of-the-art methods.

Specifically, our approach consists of two stages. 
In the first stage, we discretize whole-body motion into multiple part motions and quantize them using VQ-VAEs, providing prior knowledge of ``what is part" for the next stage. 
In the second stage, we use multiple Part-Coordinated Transformers, which are capable of communicating with each other, to generate coordinated motions of different parts. These part motions are integrated into whole-body motion subsequently.
Extensive experiments on HumanML3D~\cite{guo2022generating} and KIT-ML~\cite{plappert2016kit} demonstrate that our method can generate realistic and coordinated motions that align with the semantic descriptions.

We delineate our contributions as follows:
\begin{itemize}[leftmargin=2.2em]
\setlength{\itemsep}{0.5em}

    \item We propose ParCo, which enables the generators to better understand finer-grained parts and coordinate the generated part motions, ultimately achieving fine-grained and coordinated motion synthesis.

    \item Our approach is computationally efficient, although employing multiple part generators, and maintains excellent motion generation performance.

    \item On the HumanML3D and KIT-ML datasets, our method significantly outperforms methods that do not rely on GT motion length as our ParCo, while demonstrating comparable performance to methods that do rely on GT motion length.
    
\end{itemize}

\section{Related Work}
\label{sec:related_work}
\subsubsection{Human Motion Synthesis.} 
Tasks in the domain of human motion synthesis fall within two distinct categories: unconditional motion generation and conditional motion generation. The categorization of these tasks is based on the input signals employed.
Unconditional motion generation~\cite{yan2019convolutional,zhao2020bayesian,zhang2020perpetual, raab2023modi}, such as VPoser~\cite{pavlakos2019expressive} and ACTOR~\cite{petrovich2021action}, is a comprehensive task involving the modeling of the entire motion space, utilizing solely motion data for training and prediction.
Human motion prediction, a highly dynamic field, seeks to forecast future movements based on observed motion. Another significant domain pertains to the generation of ``in-betweening'' motions, which fill the gaps between past and future poses~\cite{duan2021single,harvey2018recurrent,harvey2020robust,kaufmann2020convolutional,tang2022real}. Unconditional motion generation commonly utilizes models well-suited for processing sequential data, including recursive~\cite{butepage2017deep,fragkiadaki2015recurrent,martinez2017human,pavllo2018quaternet}, generative adversarial~\cite{barsoum2018hp,hernandez2019human}, graph convolutional~\cite{mao2019learning}, and attention~\cite{mao2020history} approaches. This enables the efficient generation of diverse motions by concurrently processing spatial and temporal signals.
Conditional motion generation involves various multimodal data types, including text~\cite{guo2022generating, petrovich2022temos,tevet2022human,guo2022tm2t,ahuja2019language2pose,kim2023flame}, occluded pose sequences~\cite{duan2021single,harvey2020robust,tevet2022human}, images~\cite{rempe2021humor,chen2022learning}, and sound~\cite{lee2019dancing,li2022danceformer,li2021ai}. Due to the rapid advancements in NLP, text-driven human motion generation has sustained a notably active status.

\subsubsection{Text-driven Human Motion Generation.}
Text-to-motion aims to generate human motion based on input textual descriptions. In earlier studies, joint-latent models~\cite{ahuja2019language2pose,petrovich2022temos} were employed, which integrate a text encoder and a motion encoder. Text2Action~\cite{ahn2018text2action} employs a recursive model, generating motion from short texts. TEMOS~\cite{petrovich2022temos} adopts a similar approach, using self-encoding structures for both text and motion constrained by KL divergence~\cite{kullback1997information}. T2M-GPT~\cite{zhang2023t2m}, TM2T~\cite{guo2022tm2t} replace recursive encoders with transformers and GRU structure, achieving promising results. MotionCLIP~\cite{tevet2022motionclip} directly introduces the powerful zero-shot-capable CLIP~\cite{radford2021learning} text encoder, following the alignment of text and pose as in Language2Pose~\cite{ahuja2019language2pose}, and additionally renders images with CLIP image encoder for auxiliary supervision.
Subsequently, solutions based on diffusion models~\cite{chen2023executing,zhang2022motiondiffuse,shafir2023human} emerge. MDM~\cite{tevet2022human}, MotionDiffuse~\cite{zhang2022motiondiffuse},  ReMoDiffuse~\cite{zhang2023remodiffuse} introduce diffusion models based on probability mapping, enabling the generation of human motion sequences from textual descriptions. Although current research enables the convenient generation of human motions based on text, challenges persist, especially in handling complex textual descriptions involving different parts.
These approaches often treat human motion as a whole, lacking an understanding of different body parts, exhibiting limited ability of aligning text and motion.
Furthermore, other methods incorporate the concept of parts into the model to generate more granular motions~\cite{ghosh2021synthesis,zhong2023attt2m,siyao2022bailando}. However, these approaches encounter challenges, including the lack of coordination among different part motions and difficulties for networks to comprehend part concepts. In contrast, our ParCo demonstrates a superior understanding of parts and effectively coordinates part motions, with lower computational complexity.

\section{Method}
\label{sec:method}

Our method consists of two stages to generate motion with an understanding of part motions.
In the first stage, we discretize the whole-body motion into multiple part motions to provide prior knowledge of ``what is part" for the second stage. 
In the second stage, the objective is to enable the model to learn the concept of part and achieve mutual coordination among multiple part motion generators.
With this design, our method can handle textual inputs involving different parts and generate human motion that aligns with the semantic descriptions in the text.

\begin{figure*}[t]
    \centering
    \includegraphics[width=\linewidth]{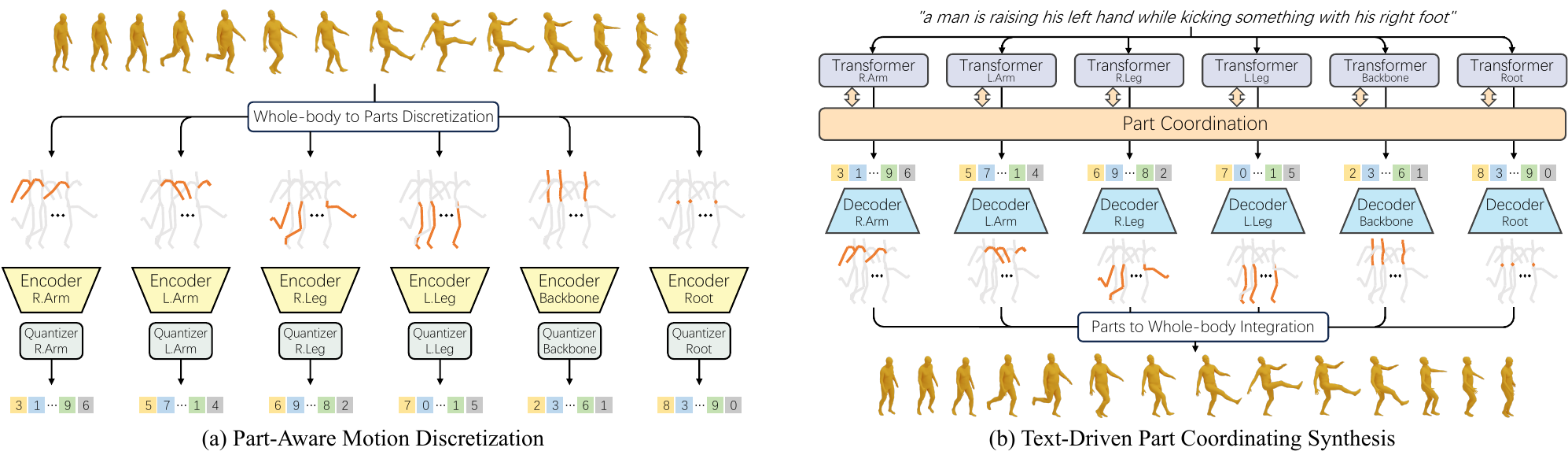}
    \caption{
    Pipeline of ParCo.
    ParCo consists of two stages: \textbf{(a)} The whole-body motion is discretized into 6 part motions, and encoded into 6 quantized code index sequences by 6 VQ-VAEs (encoder and quantizer). This process provides a priori about the concept of part motions for the second stage.
    \textbf{(b)} We use the quantized index sequences and corresponding textual description to train 6 transformers for part motion generation. At the same time, these generators are coordinated by our Part Coordination module. The generated part motion codes are decoded by VQ-VAE (decoder) to reconstruct the 6 part motions, which will be integrated into the final whole-body motion.
    }
    \label{fig:method_pipline}
\end{figure*}

\subsection{Part-Aware Motion Discretization}
In stage 1, our method partitions the whole-body motion into multiple part motions and independently encodes each of these part motions using a VQ-VAE.
This ensures that each part motion possesses an independent representation space (encoding space), providing prior knowledge about the concept of part motion for the next stage. 

The 3D human body models (e.g. SMPL, MMM) typically use Kinematic Trees to model human skeleton as 5 chains (limbs and backbone) for motion modeling. We inherit this division and add the Root part to represent trajectories.
Therefore, we divide the whole-body motion into six parts motions: R.Arm, L.Arm, R.Leg, L.Leg, Backbone, and Root.
As illustrated in Fig.~\ref{fig:method_pipline} (a), the first four represent the right/left arm and right/left leg, while the backbone denotes the spine and skull, and the root represents the pelvis joint's movement information.
The commonly used human motion datasets, HumanML3D~\cite{guo2022generating} and KIT-ML~\cite{plappert2016kit}, utilize different body skeleton models, SMPL~\cite{loper2023smpl} and MMM~\cite{terlemez2014master} respectively. We provide a detailed explanation of how we perform our six-part partitioning for these two skeleton models in the supplementary material.

The aforementioned partitioning process can be formalized as:
given a motion sequence $M=\left[ m_1, ..., m_C \right]$, where C is the number of frames, we separate it into part motions $\left\{ P^i=\left[ p^i_1, ..., p^i_C\right] \right\}, i\in\left[ 1,...,S\right]$, where $S$ is the number of parts and $m_*=[p^1_*,...,p^S_*]$.

After separating the whole-body motion into part motions,
we further discretize the part motions into code sequences using VQ-VAE. We utilize this discretized representation in the next stage's generation process, as it exhibits better generalization capabilities and contributes to the improvement of training and inference efficiency. 

Firstly, we use ${Encoder}^i$ to get the $i$-th part motion's encoding that $E^i=\\{Encoder}^i\left(P^i\right)=\left[ e^i_1, ..., e^i_l, ..., e^i_L\right]$, where $L=\frac{C}{r}$ and $r$ is the downsampling rate of encoder. 
Then, we discretize the $E^i$ into $Q^i=[ v^i_{k^i_1}, ..., v^i_{k^i_l}, ..., v^i_{k_L} ]$ according to a learnable codebook $V^i=\left\{ v^i_j\right\}, j=1,...,J$, where $J$ is the number of codes in the codebook.
The index $k_l$ is obtained by finding the most similar code:
\begin{equation}\label{eq:quantization}
\begin{aligned}
k^i_l&=\mathop{\arg\min}\limits_{j\in \left\{ 1, ..., J\right\}} \left\| e^i_l - v^i_j \right\|.
\end{aligned}
\end{equation}
By this way, we discretize the motion $M$ into $S$ discretized part motion representations $\{Q^i\}, i=1,...,S$.

For training the VQ-VAE, We use a decoder to reconstruct the $i$-th part motion $\hat{P^i}={Decoder}^i\left( Q^i \right)$ with reconstruction loss $\mathcal{L}^i_r=\| \hat{P^i}-P^i \|$.
The optimization objective of the $i$-th part's VQ-VAE is:
\begin{equation}\label{eq:loss_VQVAE}
\begin{aligned}
    \mathcal{L}^i= \mathcal{L}^i_r + \| sg(E^i) - Q^i\| +\beta\| E^i - sg(Q^i)\|,
\end{aligned}
\end{equation}
where $sg$ represents the stop-gradient operation. 
The first term is the reconstruction loss function, ensuring that the VQ-VAE can reconstruct the original part motion from the encoding. 
The second term is the codebook loss function.
And the third term is the commitment loss, aiming to make the representation $e^i_l$ output by the encoder as close as possible to the code $v^i_{k_l}$ contained in the codebook. The weight of this loss is controlled by the hyperparameter $\beta$.

\subsection{Text-Driven Part Coordination}

\begin{figure}[!t]
   \centering
   \includegraphics[width=0.70\linewidth, trim={0 0 0.3cm 0},clip]{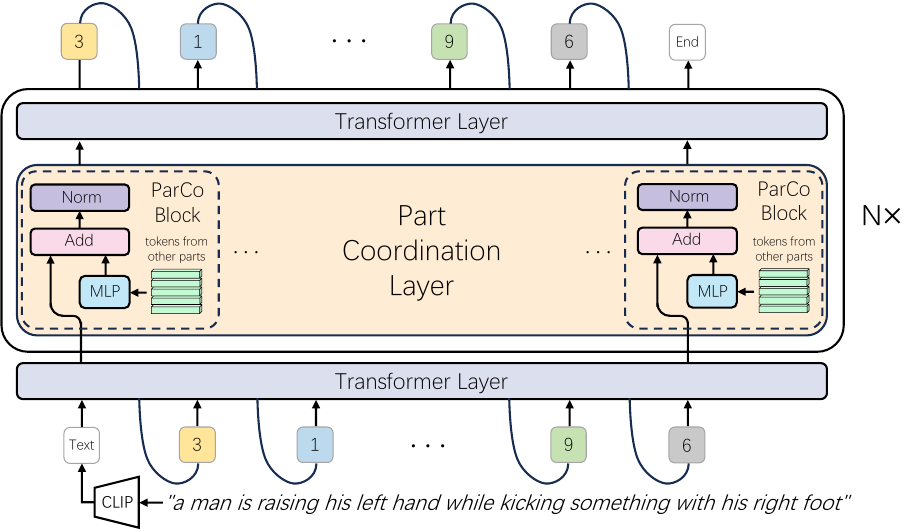}
   \caption{
   The architecture of our Part-Coordinated Transformer.
   }
   \label{fig:method_part_coord_transformer}
\end{figure}
In this stage, we employ transformers as generators to achieve text-to-motion generation. 
Diverging from the approach of using a large transformer for the entire whole-body motion, we utilize multiple small transformers to generate each part motion code sequence obtained from the previous stage. 
This allows small transformers to be aware of the conception of ``part motion'' constructed before. 
However, relying solely on these separate part motion generators can lead to an inability to collectively generate whole-body motions, due to a lack of knowledge about the motions of other parts. 
To address this, we introduce the Part Coordination module, facilitating communication among all part transformer generators to collaboratively generate whole-body motion.
This capability enables our ParCo to handle textual inputs involving different parts, generating human motion that aligns with semantic descriptions in the text.

As depicted in Fig.~\ref{fig:method_pipline} (b), we employ six transformers to generate code index sequences of part motions based on the input text. At the same time, these six transformers collaborate through our Part Coordination module to coordinate with each other. Subsequently, these generated index sequences are decoded into the original representation of part motions by the decoder of the VQ-VAE. These part motions are integrated to form whole-body motion.

Specifically, we model the whole text-to-motion generation process as estimating the distribution $p(M|t)$ of motion $M$ given the text $t$.
Since we have discretized the whole-body motion into part motions, we can model the entire motion distribution $p(M|t)$ through the estimation of conditional distributions $p(K^i|t)$ for each part, where $\{ K^i=[k^i_1,...,k^i_L]\}, i\in[1,...,S]$ is the code indices obtained in the first stage.

To model $i$-th part motion's distribution, we propose an autoregressive distribution, which allows parts to coordinate with each other,
\begin{equation}\label{eq:part_coord_autoreg_dist}
\begin{aligned}
    p(K^i|t) & = \prod\limits_{h=1}^L p(k^i_h|k^i_1, o^i_1, ;...; k^i_{h-1}, o^i_{h-1};t), \\
             o^i_{*} & =  \left\{ k^j_* \right\}, {j \neq i, j \in [1,...,S]}.\\
\end{aligned}
\end{equation}
When predicting token $k^i_h$ for the $i$-th part, the prediction not only relies on all tokens predicted by itself from time $1$ to $h-1$ but is also conditioned by predictions from all other parts during the same time span. Also, after the $i$-th part transformer predicts the token $K^i_h$, the prediction will be used by all part generators to predict the next token.

Finally, we learn the entire body motion by estimating the distributions of all part motions,
\begin{equation}\label{eq:loss_transformers}
\begin{aligned}
    \mathcal{L}&=\mathbb{E}_{M,t \sim p(M,t)} [-\log p(M|t)]\\
  &= \mathbb{E}_{M,t\sim p(M,t)} [- \sum\limits_{i=1}^S \log p(K^i|t)].
\end{aligned}
\end{equation}

We further propose Part-Coordinated Transformer, a transformer capable of coordinating with other transformers, to approximate $p(K^i|t)$.
As shown in Fig.~\ref{fig:method_part_coord_transformer}, we insert a Part Coordination Layer before each transformer layer (except for the first transformer layer).
For each token $x^i$ output by the previous transformer layer, it passes through our ParCo Block.
The ParCo Block coordinates with other part motion generators, fusing current token $x^i$ with tokens from other part transformers,
\begin{equation}\label{eq:ParCo_Block}
\begin{aligned}
    x^i_{coord}&=LN(x^i+MLP^i(y)), \\
    \ \ y&=\left\{ x^j \right\}, {j \neq i, \ j \in [1,...,S]},
\end{aligned}
\end{equation}
where $LN$ denotes the LayerNorm operation, and $y$ represents the tokens from other transformers' layers. The fused token $x^i_{coord}$ is then input into the subsequent transformer layer.

\subsection{Discussion}

Text-to-motion synthesis intrinsically consists of two stages, i.e., body part representation and generation, to fulfill actions described by texts. To precisely ground each word to the corresponding body part, Balando~\cite{siyao2022bailando} partitions the body into upper and lower parts for motion quantization while reconstructing through a shared decoder. SCA~\cite{ghosh2021synthesis} steps further to equip independent generators for flexible correspondence. However, coarse-grained sub-body level modeling yields sub-optimal results. 

AttT2M~\cite{zhong2023attt2m} introduces a global-local attention mechanism to learn hierarchical body-part semantics for accurate motion synthesis. SINC~\cite{athanasiou2023sinc} adopts a simple additive composition of part motion for GPT-guided synthetic training data creation. In contrast, we advocate explicitly discretizing body parts as individual action atoms and synthesis motion with decentralized generators and a centralized part coordinating module. With a series of computation-economic designs, we keep the relative independence yet close coordination relationships of body parts and report superior results.


\section{Experiment}
\label{sec:experiment}


\begin{table}[ht]
  \caption{
    Comparisons to current state-of-the-art methods on HumanML3D test set.
    ``$\uparrow$'' denotes that higher is better.
    ``$\downarrow$'' denotes that lower is better.
    ``$\rightarrow$'' denotes that results are better if the metric is closer to the real motion.
    \textbf{Bold} and \underline{underlined} indicate the best and second-best results, respectively. \S~reports results using ground-truth motion length. The results of ReMoDiffuse* are obtained from official checkpoints and employ uniform random sampling of motion lengths as input.}
  \centering
  \resizebox{1.\linewidth}{!}
{
  \begin{tabular}{lccccccc}
    \toprule[1.25pt]
\multirow{2}{*}{{Methods}} &\multicolumn{3}{c}{{R-Precision $\uparrow$}} & \multirow{2}{*}{{FID $\downarrow$}} & \multirow{2}{*}{{MM-Dist $\downarrow$}} & \multirow{2}{*}{{Diversity $\rightarrow$}} & \multirow{2}{*}{{MModality $\uparrow$}}\\
\cmidrule(rl){2-4}
  & Top-1 & Top-2 & Top-3 \\ 
    \midrule
    Real motion & $0.511^{\pm.003}$ & $0.703^{\pm.003}$ & $0.797^{\pm.002}$ & $0.002^{\pm.000}$ & $2.974^{\pm.008}$ & $9.503^{\pm.065}$ & - \\ 
    \midrule
    MDM$^{\S}$~\cite{tevet2022human} & $0.320^{\pm.005}$ & $0.498^{\pm.004}$ & $0.611^{\pm.007}$ & $0.544^{\pm.044}$ & $5.566^{\pm.027}$ & ${9.559^{\pm.086}}$ & ${2.799^{\pm.072}}$ \\
    MLD$^{\S}$~\cite{chen2023executing} & $0.481^{\pm.003}$ & $0.673^{\pm.003}$ & $0.772^{\pm.002}$ & $0.473^{\pm.013}$ & $3.196^{\pm.010}$ & $9.724^{\pm.082}$ & $2.413^{\pm.079}$ \\
    MotionDiffuse$^{\S}$~\cite{zhang2022motiondiffuse} & $0.491^{\pm.001}$ & ${0.681^{\pm.001}}$ & ${0.782^{\pm.001}}$ & $0.630^{\pm.001}$ & $3.113^{\pm.001}$ & $9.410^{\pm.049}$ & $1.553^{\pm.042}$ \\
    ReMoDiffuse$^{\S}$~\cite{zhang2023remodiffuse} & ${0.510^{\pm.005}}$ & ${0.698^{\pm.006}}$ & ${0.795^{\pm.004}}$ & ${{0.103^{\pm.004}}}$ & ${2.974^{\pm.016}}$ & $9.018^{\pm.075}$ & $1.795^{\pm.043}$ \\
    ReMoDiffuse$^{*}$ & ${0.450^{\pm.003}}$ & $0.638^{\pm.002}$ & $0.743^{\pm.003}$ & ${0.281^{\pm.010}}$ & $3.271^{\pm.008}$ & $9.236^{\pm.085}$ & - \\
    \midrule
    Text2Gesture~\cite{bhattacharya2021text2gestures} & $0.165^{\pm.001}$ & $0.267^{\pm.002}$ & $0.345^{\pm.002}$ & $7.664^{\pm.030}$ & $6.030^{\pm.008}$ & $6.409^{\pm.071}$ & - \\
    Seq2Seq~\cite{plappert2018learning} & $0.180^{\pm.002}$ & $0.300^{\pm.002}$ & $0.396^{\pm.002}$ & $11.75^{\pm.035}$ & $5.529^{\pm.007}$ & $6.223^{\pm.061}$ & - \\
    Language2Pose~\cite{ahuja2019language2pose} & $0.246^{\pm.001}$ & $0.387^{\pm.002}$ & $0.486^{\pm.002}$ & $11.02^{\pm.046}$ & $5.296^{\pm.008}$ & $7.676^{\pm.058}$ & - \\
    Hier~\cite{ghosh2021synthesis} & $0.301^{\pm.002}$ & $0.425^{\pm.002}$ & $0.552^{\pm.004}$ & $6.532^{\pm.024}$ & $5.012^{\pm.018}$ & $8.332^{\pm.042}$ & - \\
    TEMOS~\cite{petrovich2022temos} & $0.424^{\pm.002}$ & $0.612^{\pm.002}$ & $0.722^{\pm.002}$ & $3.734^{\pm.028}$ & $3.703^{\pm.008}$ & $8.973^{\pm.071}$ & $0.368^{\pm.018}$ \\
    TM2T~\cite{guo2022tm2t} & $0.424^{\pm.003}$ & $0.618^{\pm.003}$ & $0.729^{\pm.002}$ & $1.501^{\pm.017}$ & $3.467^{\pm.011}$ & $8.589^{\pm.076}$ & $\underline{2.424^{\pm.093}}$ \\
    T2M~\cite{guo2022generating} & $0.457^{\pm.002}$ & $0.639^{\pm.003}$ & $0.740^{\pm.003}$ & $1.067^{\pm.002}$ & $3.340^{\pm.008}$ & $9.188^{\pm.002}$ & $2.090^{\pm.083}$ \\
    T2M-GPT~\cite{zhang2023t2m} & ${0.492^{\pm.003}}$ & $0.679^{\pm.002}$ & $0.775^{\pm.002}$ & ${0.141^{\pm.005}}$ & $3.121^{\pm.009}$ & $9.722^{\pm.082}$ & $1.831^{\pm.048}$ \\
    Fg-T2M~\cite{wang2023fg} & ${0.492^{\pm.002}}$ & $0.683^{\pm.003}$ & $0.783^{\pm.002}$ & ${0.243^{\pm.019}}$ & $3.109^{\pm.007}$ & $9.278^{\pm.072}$ & $1.614^{\pm.049}$ \\
    AttT2M~\cite{zhong2023attt2m} & $\underline{0.499^{\pm.003}}$ & $\underline{0.690^{\pm.002}}$ & $\underline{0.786^{\pm.002}}$ & $\underline{0.112^{\pm.006}}$ & $\underline{3.038^{\pm.007}}$ & $\underline{9.700^{\pm.090}}$ & $\bm{2.452^{\pm.051}}$ \\
    \midrule
    \rowcolor{aliceblue!60} ParCo (Ours) & $\bm{0.515^{\pm.003}}$ & $\bm{0.706^{\pm.003}}$ & $\bm{0.801^{\pm.002}}$ & ${\bm{0.109^{\pm.005}}}$ & $\bm{2.927^{\pm.008}}$ & ${\bm{9.576^{\pm.088}}}$ & ${1.382^{\pm.060}}$\\
  \bottomrule
  \label{tab:humanml3d}
  \end{tabular}}
\end{table}


\begin{table}[ht]
  \caption{
    Comparisons to current state-of-the-art methods on KIT-ML test set.
}
  \centering
  
\resizebox{1.\linewidth}{!}{
  \begin{tabular}{lccccccc}
    \toprule[1.25pt]
\multirow{2}{*}{{Methods}} &\multicolumn{3}{c}{{R-Precision $\uparrow$}} & \multirow{2}{*}{{FID $\downarrow$}} & \multirow{2}{*}{{MM-Dist $\downarrow$}} & \multirow{2}{*}{{Diversity $\rightarrow$}} & \multirow{2}{*}{{MModality $\uparrow$}}\\
\cmidrule(rl){2-4}
  & Top-1 & Top-2 & Top-3 \\ 
    \midrule
    Real motion & $0.424^{\pm.005}$ & $0.649^{\pm.006}$ & $0.779^{\pm.006}$ & $0.031^{\pm.004}$ & $2.788^{\pm.012}$ & $11.08^{\pm.097}$ & - \\ 
    \midrule
    MDM$^{\S}$~\cite{tevet2022human} & $0.164^{\pm.004}$ & $0.291^{\pm.004}$ & $0.396^{\pm.004}$ & $0.497^{\pm.021}$ & $9.191^{\pm.022}$ & $10.85^{\pm.109}$ & $1.907^{\pm.214}$ \\
    MLD$^{\S}$~\cite{chen2023executing} & $0.390^{\pm.008}$ & $0.609^{\pm.008}$ & $0.734^{\pm.007}$ & ${0.404^{\pm.027}}$ & $3.204^{\pm.027}$ & $10.80^{\pm.117}$ & $2.192^{\pm.071}$ \\
    MotionDiffuse$^{\S}$~\cite{zhang2022motiondiffuse} & $0.417^{\pm.004}$ & $0.621^{\pm.004}$ & $0.739^{\pm.004}$ & $1.954^{\pm.062}$ & $2.958^{\pm.005}$ & ${11.10^{\pm.143}}$ & $0.730^{\pm.013}$ \\
    ReMoDiffuse$^{\S}$~\cite{zhang2023remodiffuse} & ${{0.427^{\pm.014}}}$ & ${0.641^{\pm.004}}$ & ${0.765^{\pm.055}}$ & ${{0.155^{\pm.006}}}$ & ${2.814^{\pm.012}}$ & $10.80^{\pm.105}$ & $1.239^{\pm.028}$ \\
    ReMoDiffuse$^{*}$ & ${0.382^{\pm.005}}$ & $0.586^{\pm.007}$ & $0.706^{\pm.006}$ & ${0.589^{\pm.022}}$ & $3.324^{\pm.030}$ & $10.31^{\pm.065}$ & - \\
    \midrule
    Seq2Seq~\cite{plappert2018learning} & $0.103^{\pm.003}$ & $0.178^{\pm.005}$ & $0.241^{\pm.006}$ & $24.86^{\pm.348}$ & $7.960^{\pm.031}$ & $6.744^{\pm.106}$ & - \\
    Text2Gesture~\cite{bhattacharya2021text2gestures} & $0.156^{\pm.004}$ & $0.255^{\pm.004}$ & $0.338^{\pm.005}$ & $12.12^{\pm.183}$ & $6.946^{\pm.029}$ & $9.334^{\pm.079}$ & - \\
    Language2Pose~\cite{ahuja2019language2pose} & $0.221^{\pm.005}$ & $0.373^{\pm.004}$ & $0.483^{\pm.005}$ & $6.545^{\pm.072}$ & $5.147^{\pm.030}$ & $9.073^{\pm.100}$ & - \\
    Hier~\cite{ghosh2021synthesis} & $0.255^{\pm.006}$ & $0.432^{\pm.007}$ & $0.531^{\pm.007}$ & $5.203^{\pm.107}$ & $4.986^{\pm.027}$ & $9.563^{\pm.072}$ & $2.090^{\pm.083}$ \\
    TM2T~\cite{guo2022tm2t} & $0.280^{\pm.005}$ & $0.463^{\pm.006}$ & $0.587^{\pm.005}$ & $3.599^{\pm.153}$ & $4.591^{\pm.026}$ & $9.473^{\pm.117}$ & $\bm{3.292^{\pm.081}}$ \\
    TEMOS~\cite{petrovich2022temos} & $0.353^{\pm.006}$ & $0.561^{\pm.007}$ & $0.687^{\pm.005}$ & $3.717^{\pm.051}$ & $3.417^{\pm.019}$ & $10.84^{\pm.100}$ & $0.532^{\pm.034}$ \\
    T2M~\cite{guo2022generating} & $0.370^{\pm.005}$ & $0.569^{\pm.007}$ & $0.693^{\pm.007}$ & $2.770^{\pm.109}$ & $3.401^{\pm.008}$ & $10.91^{\pm.119}$ & $1.482^{\pm.065}$ \\
    AttT2M~\cite{zhong2023attt2m} & ${0.413^{\pm.006}}$ & $\underline{0.632^{\pm.006}}$ & $\underline{0.751^{\pm.006}}$ & ${0.870^{\pm.039}}$ & $3.039^{\pm.021}$ & $\bm{10.96^{\pm.123}}$ & $\underline{2.281^{\pm.047}}$ \\
    T2M-GPT~\cite{zhang2023t2m} & $0.416^{\pm.006}$ & $0.627^{\pm.006}$ & $0.745^{\pm.006}$ & $\underline{0.514^{\pm.029}}$ & $\underline{3.007^{\pm.023}}$ & $10.92^{\pm.108}$ & $1.570^{\pm.039}$ \\ 
    Fg-T2M~\cite{wang2023fg} & $\underline{0.418^{\pm.005}}$ & $0.626^{\pm.004}$ & $0.745^{\pm.004}$ & ${0.571^{\pm.047}}$ & $3.114^{\pm.015}$ & $10.93^{\pm.083}$ & $1.019^{\pm.029}$ \\
    \midrule
    \rowcolor{aliceblue!60} ParCo (Ours) & $\bm{0.430^{\pm.004}}$ & $\bm{0.649^{\pm.007}}$ & $\bm{0.772^{\pm.006}}$ & $\bm{0.453^{\pm.027}}$ & ${\bm{2.820^{\pm.028}}}$ & $\underline{10.95^{\pm.094}}$ & $1.245^{\pm.022}$\\
  \bottomrule
  \label{tab:kit}
  \end{tabular}}
\end{table}

\subsection{Settings}

\subsubsection{Datasets.} 
We utilized two widely used text-to-motion datasets, KIT-ML and HumanML3D, for training and testing our method. Comparative evaluations were conducted on these datasets against other existing methods.
Processed from KIT~\cite{plappert2016kit} and CMU~\cite{cmu2016mocap} datasets, KIT-ML Dataset~\cite{plappert2016kit} comprises 3,911 sequences of human body motions with 6,278 text annotations.  Each motion is annotated with 1 to 4 text descriptions, averaging approximately 8 words each. KIT-ML employs the MMM~\cite{terlemez2014master} skeletal model which has 21 joints, and we detail in the supplementary materials how we partition its joints into 6 parts. We follow the train, validation, and test set divisions as outlined in~\cite{guo2022generating} and report our ParCo's performance on the test set.
HumanML3D~\cite{guo2022generating} Dataset is the largest dataset with 14,616 3D human body motion sequences and 44,970 corresponding textual descriptions, sourced from AMASS~\cite{mahmood2019amass} and HumanAct12~\cite{guo2020action2motion}. Each motion includes a minimum of 3 text descriptions, with an average length of 12 words. HumanML3D adopts the SMPL~\cite{loper2023smpl} skeletal model which has 22 joints, and we detail in the supplementary materials how we partition its joints into 6 parts. Similar to KIT-ML, we follow the division into train, validation, and test sets as specified in~\cite{guo2022generating} and report our ParCo's performance on the test set.

\subsubsection{Evaluation Metrics.}
Following prior text-to-motion work, we leverage pre-trained text and motion feature extractors~\cite{guo2022generating} to measure cross-modal alignment and semantic similarity between motions, rather than joint coordinate position. Based on the extracted features, we employ the following evaluation metrics:
\textbf{(i) R-Precision}: Reflects the accurate semantic matching between text and motion. We calculate Top-1, Top-2, and Top-3 accuracy based on the Euclidean distance between a given motion sequence and 32 text descriptions (1 ground truth and 31 randomly selected non-matching).
\textbf{(ii) FID}: We use FID~\cite{heusel2017gans} to quantify the distributional disparity between generated and real motions based on extracted motion features. It's crucial to highlight that FID does not assess the alignment between textual descriptions and generated motions.
\textbf{(iii) MM-Dist}: Measures the Euclidean distance between feature vectors of text and motion, reflecting the semantic similarity.
\textbf{(iv) Diversity}: Indicates the variance in generated motions. We randomly sample two equal-sized subsets from all motions, calculating the average Euclidean distance between the subsets. Closer diversity values between generated and real motions signify a better match.
\textbf{(v) MModality}: Reflects the diversity of generated motions for a given text. We generate 10 pairs of motions for each text, compute feature vectors' distance between each pair, and take the average. Given that incorrectly generated motions lead to a high MModality value, it fails to reflect the alignment between motions and texts.
In addition, following~\cite{tevet2022human}, we run each evaluation 20 times (except MModality for 5 times) and report the average with a 95\% confidence interval.


\begin{figure*}[ht]
    \centering
    \includegraphics[width=\linewidth]{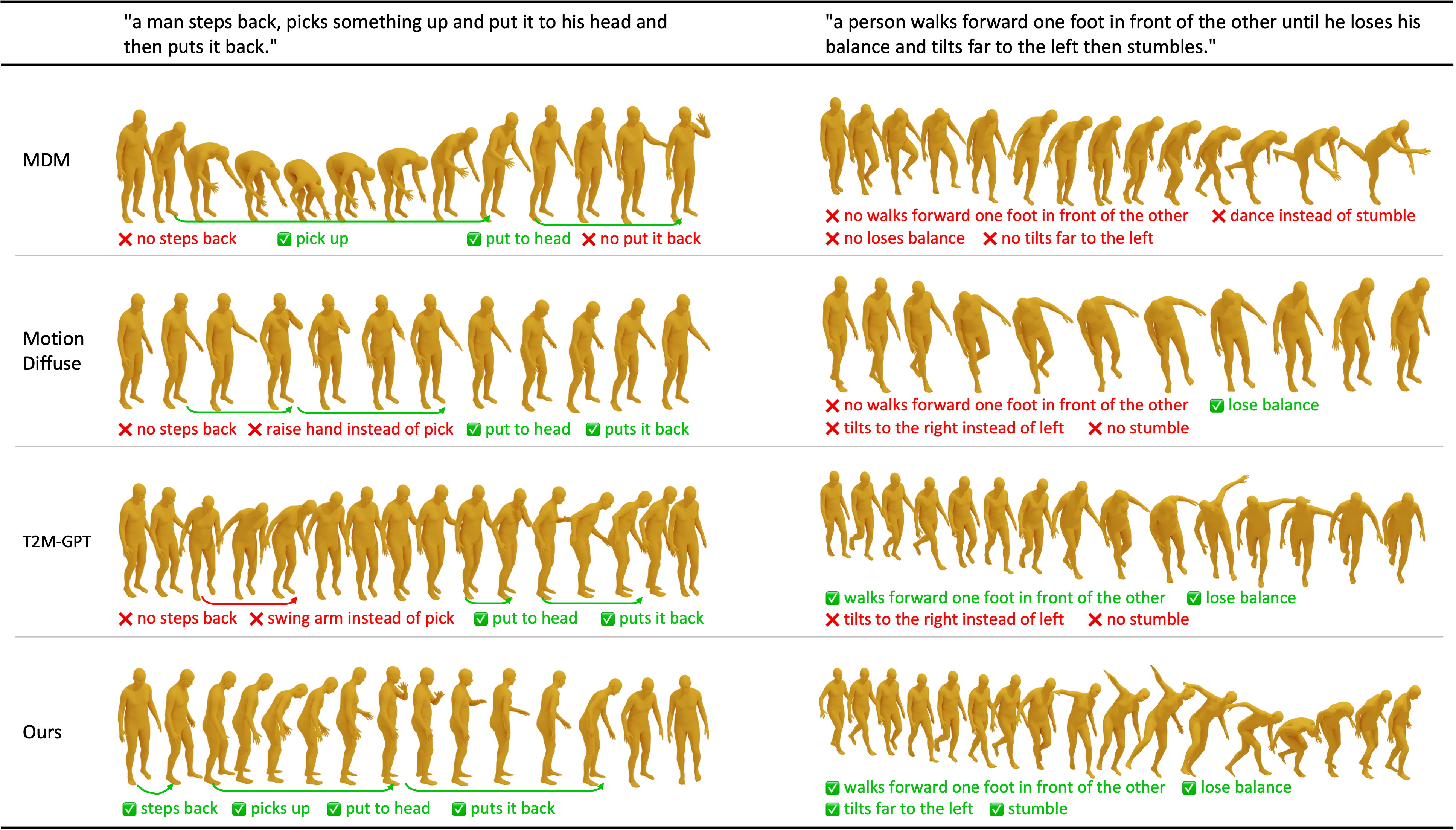}
    \caption{
    Qualitative comparison with existing methods.
    \textcolor[RGB]{0,196,12}{Green} indicates the motion is consistent with the text description.
    \textcolor{red}{Red} indicates the text description lacks the corresponding motion or got the wrong motion.
    }
    \label{fig:fig_exp_qualitative_comparison}
\end{figure*}

\subsubsection{Implementation Details.}
Our ParCo employs 6 small VQ-VAEs for discretizing part motions and 6 small transformers with Part Coordination modules for text-to-motion generation. For VQ-VAEs, all codebooks contain 512 codes, and all parts have 128 code dimensions, except for the Root part, which has a code dimension of 64. The encoder's downsampling rate is set to r=4. For transformers, each has 14 layers, and token dimension is 256.
And we insert a Part Coordination Layer before all remaining layers except for the first transformer layer. 
ParCo Blocks in the same layer of the same transformer share parameter weights.
We set the number of MLP layers in ParCo Block to 3.
For training VQ-VAE, we use a learning rate of 2e-4 before 200K and 1e-5 after 200K, AdamW~\cite{loshchilov2017decoupled} optimizer with $beta_1=0.9$ and $beta_2=0.99$, and batch size of 256. The commitment loss weight $\beta$ is set to 1.0. 
For training transformer, we use a learning rate of 1e-4 before 150K and 5e-6 after 150K, AdamW optimizer with $beta_1=0.5$ and $beta_2=0.99$, and batch size of 128.

\subsection{Comparisons to State-of-the-art}
We compare our ParCo with other methods (including the methods using ground-truth motion length) on HumanML3D test set (Table.~\ref{tab:humanml3d}) and KIT-ML test set (Table.~\ref{tab:kit}).
Our method demonstrates superior performance compared to previous state-of-the-art methods on R-Precision and MM-Dist, and comparable results on FID, indicating ParCo's superiority.
As for Top-1, Top-2, and Top-3 of R-Precision, our ParCo surpasses previous SOTA, ReMoDiffuse~\cite{zhang2023remodiffuse} (using GT motion length), with 0.005, 0.008, and 0.006 on HumanML3D, 0.003, 0.008, and 0.007 on KIT-ML.
And for MM-Dist, our ParCo exceeds ReMoDiffuse's performance with 0.047 on HumanML3D and has comparable result on KIT-ML.
In terms of FID, our result is on par with ReMoDiffuse on HumanML3D, and achieve the SOTA performance on HumanML3D and KIT-ML compared to the methods not relying GT information.
Furthermore, our ParCo yields a lower MModality value compared to the previous state-of-the-art. 
This may be attributed to our ParCo generating motions that align more accurately with the text, thereby reducing the occurrence of irrelevant or incorrect motions.
Qualitative results presented in Fig.~\ref{fig:fig_exp_qualitative_comparison} demonstrate that our generated motions are more realistic, coordinated, and aligned with textual descriptions for texts involving multiple body parts.
Details such as "steps back," "tilts far to the left," and "stumbles" are accurately captured by our ParCo, while other methods either ignore or incorrectly generate these nuanced actions.
Besides, as illustrated in Fig.~\ref{tab:humanml3d} and Fig.~\ref{tab:humanml3d_trainvaltest_analysis}, ParCo is endowed the part-level composition superiority across action-level spatial and temporal compositions respectively.
More details are provided in the supplementary materials.

\subsubsection{GT Leakage of Diffusion-based Methods}

Previous studies~\cite{tevet2022human,chen2023executing,zhang2022motiondiffuse,zhang2023remodiffuse} based on the Diffusion model utilize the ground truth motion length as an input for synthesizing motions during evaluation, contributing to their remarkable FID scores.
However, this approach is impractical for real-world applications.
For clarity, we replace the GT motion lengths with random lengths sampled uniformly from dataset's motion length range to evaluate ReMoDiff.
On HumanML3D (KIT-ML), the ReMoDiff's Top-1, Top-2, and Top-3 R-precision decrease by 0.041 (0.053), 0.040 (0.068), and 0.033 (0.075) respectively, while the FID increases by 0.147 (0.426), justifying our speculation. 
As an auto-regressive approach, our ParCo demonstrates competitive R-Precision and FID without requiring pre-defined motion lengths Table~\ref{tab:humanml3d}.
Given its lower computational and parameter consumption (Table.~\ref{tab:param_size_gen_compare}), the superiority of ParCo is convincing.

\subsection{Analysis}

\subsubsection{Performance on Text Inputs of Different Lengths.}
\begin{figure*}[t]
    \centering
    \includegraphics[width=\linewidth]{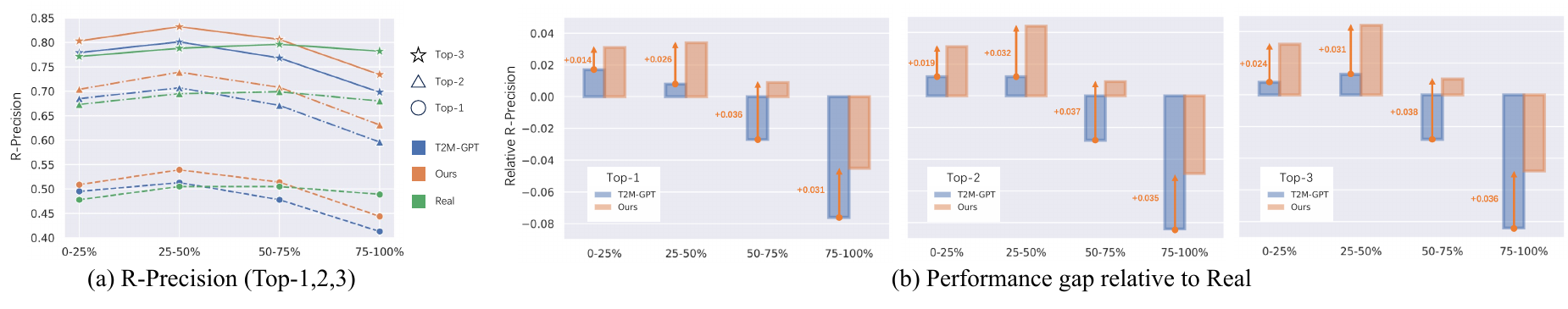}
    \caption{
    Comparison on 4 HumanML3D test subsets divided based on text length. 
    0-25\%, 25-50\%, 50-75\%, and 75-100\% respectively represent four subsets based on text length from short to long.
    }
    \label{fig:all_and_relative_RPrecision}
\end{figure*}

In order to investigate the synthetic performance given textual descriptions of different lengths, we sort the HumanML3D test set based on the length of textual descriptions, and split it into four subsets with approximately equal numbers of text-motion pairs: 0-25\%, 25-50\%, 50-75\%, and 75-100\%, from short to long. 
The details of these splits are available in the supplementary materials.
We conducted evaluations of real motion, T2M-GPT~\cite{zhang2023t2m}, and our method on these four splits.
As illustrated in Fig.~\ref{fig:all_and_relative_RPrecision}, our method exhibits comprehensive improvements in all subsets when contrasted with the baseline~\cite{zhang2023t2m}.
It demonstrates that our design is beneficial for text inputs of different lengths without compromise.

\begin{table}
  \begin{minipage}{0.56\textwidth}
    \caption{
    Ablations of body discretization and part coordination module. $*$ denotes our ParCo.
}
  \centering
  \resizebox{1.\linewidth}{!}{
  \begin{tabular}{@{}r cc ccc@{}}
    \toprule[1.25pt]
~ & Discretization & Part Coord. & Top-1 $\uparrow$ & FID $\downarrow$ & MM-Dist $\downarrow$ \\
    \midrule[1.pt]
    A  & \multirow{2}{*}{{Up\&LowBody}} & $\times$ &   $0.444^{\pm.003}$ & $0.497^{\pm.015}$ & $3.367^{\pm.009}$ \\
    
    B & & \checkmark & $0.491^{\pm.003}$ & $0.172^{\pm.007}$ & $3.071^{\pm.008}$ \\
    
    \midrule[1.pt]
    C  & \multirow{2}{*}{{6 Parts}} & $\times$ &             $0.375^{\pm.003}$ & $3.652^{\pm.030}$ & $4.012^{\pm.012}$ \\
    
    *D & & \checkmark &           $0.515^{\pm.003}$ & $0.109^{\pm.005}$ & $2.927^{\pm.008}$ \\
  \bottomrule
  \label{tab:ablation}
  \end{tabular}}
  \end{minipage}%
  \hfill
  \begin{minipage}{0.4\textwidth}
  \caption{Computational complexity analysis.}
  \renewcommand{\arraystretch}{1.43}
  \centering
  \resizebox{1.\linewidth}{!}{
  \begin{tabular}{l ccc}
    \toprule[1.25pt]
Method & Param(M) & FLOPs(G) & InferTime(s) \\
    \midrule[1.25pt]
    ReMoDiff        & 198.2 & 481.0 & 0.091 \\

    T2M-GPT            & 237.6 & 292.3 & 0.544 \\
    
    ParCo   & 168.4 & 211.7 & 0.036 \\
  \bottomrule
  \label{tab:param_size_gen_compare}
  \end{tabular}}
  \end{minipage}
\end{table}


\begin{table}[t]
  \caption{Evaluation of real motion data on train, val, and test set of HumanML3D.}
  \centering
  \begin{tabular}{l ccccc}
    \toprule[1.25pt]
    Split & Top-1 & Top-2 & Top-3 & MM-Dist & Diversity \\
    \midrule[1.25pt]
    Train & $0.628^{\pm.001}$ & $0.810^{\pm.001}$ & $0.888^{\pm.001}$ & $2.388^{\pm.003}$ & $9.685^{\pm.083}$ \\

    Val   & $0.513^{\pm.004}$ & $0.703^{\pm.004}$ & $0.800^{\pm.003}$ & $2.911^{\pm.010}$ & $9.575^{\pm.081}$ \\
    
    Test  & $0.512^{\pm.003}$ & $0.703^{\pm.002}$ & $0.797^{\pm.002}$ & $2.973^{\pm.007}$ & $9.495^{\pm.079}$ \\
  \bottomrule
  \label{tab:humanml3d_trainvaltest_analysis}
  \end{tabular}
\end{table}

\subsubsection{Ablation Study.}
We investigate different discretizations for whole-body motion and conduct ablation of our Part Coordination module.
We compare our 6-part partition and upper-lower body partition proposed by SCA~\cite{ghosh2021synthesis}.
It is noteworthy that, despite SCA dividing whole-body into upper and lower body for motion generation, its generations of upper body motion and lower body motion are entirely independent and lack coordination compared with our method. 
As demonstrated in Table.~\ref{tab:ablation}, the comparison between our ParCo \textbf{(D)} and the re-implemented SCA \textbf{(A)} indicates that our method's generated motions exhibit significantly higher performance in Top-1, MM-Dist, and FID than SCA.
Furthermore, the contrasts between results \textbf{(A)} and \textbf{(B)}, as well as results \textbf{(C)} and \textbf{(D)}, underscore the necessity of our ParCo facilitating communication and coordination among different part motions during the generation process.
In addition, the comparison between results \textbf{(B)} and \textbf{(D)} validates the effectiveness of our 6-part partitioning, which makes the model aware of the concept of parts and contributes to text-to-motion synthesis.

\subsubsection{Computational complexity analysis.}
With lightweight architectures and parallel computing support, our ParCo shows superior efficiency from parameters, flops, and inference time, justifying the efficacy of our part discretization and coordination designs, Table.~\ref{tab:param_size_gen_compare}.

\begin{figure}[!t]
   \centering
   \includegraphics[width=\linewidth, trim={0 0 0.3cm 0},clip]{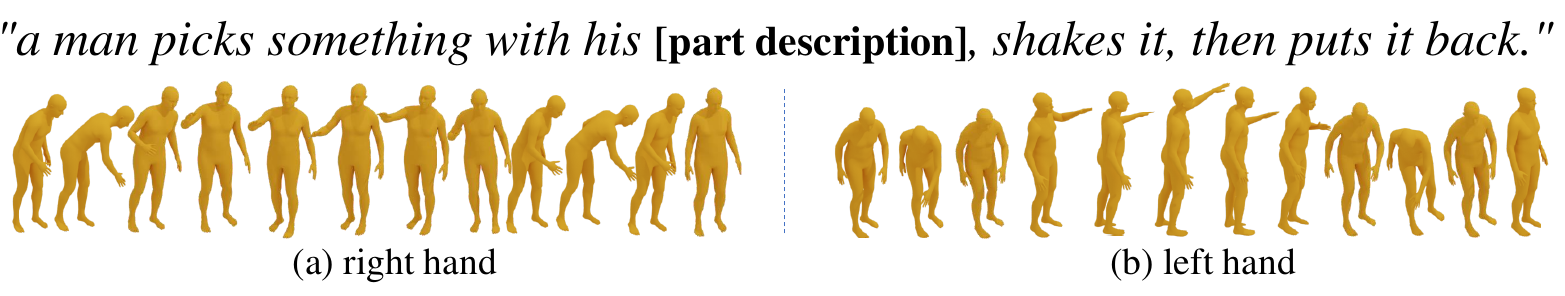}
   \caption{
   Qualitative result of left-right exchange experiment on our ParCo.
   }
   \label{fig:left_right_experiment}
\end{figure}

\subsubsection{Precise part control.}
To compare the control of different methods over the movement of human parts, we conducted a left-right exchange experiment on ten sentences. We examined several methods for generating actions and accomplishing the left-right swap task, the success rates are listed as follows: 70\% (ParCo), 50\% (T2M-GPT), 30\% (ReMoDiff), 20\% (MDM), 0\% (MoDiff). Notably, our ParCo exhibited the highest accuracy, underscoring its proficiency in the awareness of human parts.
The qualitative results are showcased in Fig.~\ref{fig:left_right_experiment}.

\subsubsection{Discussion.}
As illustrated in Table \ref{tab:humanml3d} and Table \ref{tab:kit}, our performance surpasses that of real motion due to the evaluation protocol~\cite{guo2022generating}, which uses pretrained feature extractors trained only on the train split. These extractors are more effective at extracting features from motions similar to those in the train split.
Table \ref{tab:humanml3d_trainvaltest_analysis} shows that evaluations on real motions from the Train, Val, and Test sets reveal significantly higher Top-1/2/3 and MM-Dist metrics on the Train split. This indicates better extractor understanding of the Train split data.
Our model, trained on the train split, generates motions resembling the train data, making feature extraction and matching easier, leading to superior performance on the test set.
Due to the lack of a comprehensive metric for text-to-motion semantic alignment, motion generation fidelity, and diversity, we advocate for a more holistic evaluation within the community.


\section{Conclusion}
\label{sec:conclusion}
In this study, we focus on enhancing the text-to-motion generation model's ability to comprehend part concepts and facilitate communication between different part motion generators, ultimately yielding the synthesis of coordinated and fined-grained motion.
Specifically, we discretize whole-body motion into multiple-part motions to establish the prior concept of parts. 
Afterward, we employ multiple lightweight generators designed to synthesize different part motions and coordinate them through our part coordination module. 
Extensive experiments showcase that our method achieves higher consistency between generated motions and textual descriptions compared to previous SOTA methods. Furthermore, in-depth analytical results suggest that our approach excels in achieving more precise part control and has lower computational complexity.
More encouragingly, our method exhibits adaptability to various part separation schemes and holds the potential for further refinement toward hierarchical part motion. We anticipate that this will have a far-reaching impact on the community.

\section*{Acknowledgements}
This work was supported by the National Key R\&D Program of China under Grant 2018AAA0102801.

%
%
\bibliographystyle{splncs04}
\bibliography{main}

\input{suppl}

\end{document}

%% file: suppl.tex







\newpage

\appendix


    
    
    

    

\section{Whole-body to Part Motions Discretization}
\label{sec:whole2parts}
The HumanML3D~\cite{guo2022generating} and KIT-ML~\cite{plappert2016kit} datasets utilize the SMPL~\cite{loper2023smpl} and MMM~\cite{terlemez2014master} Human Models, respectively. These datasets include joints related to whole-body motion, excluding hand joints, as depicted in Fig.~\ref{fig:suppl_smpl_division} and Fig.~\ref{fig:suppl_mmm_division}. 
In addition, HumanML3D utilizes 22 joints from the SMPL human model, while the widely-used preprocessed KIT-ML benchmark, provided by~\cite{guo2022generating}, comprises 21 joints.

\subsubsection{ParCo's 6-Part Division}
\begin{figure}[t]
    \centering
    \begin{minipage}{0.48\textwidth}
        \centering
        \includegraphics[width=\linewidth]{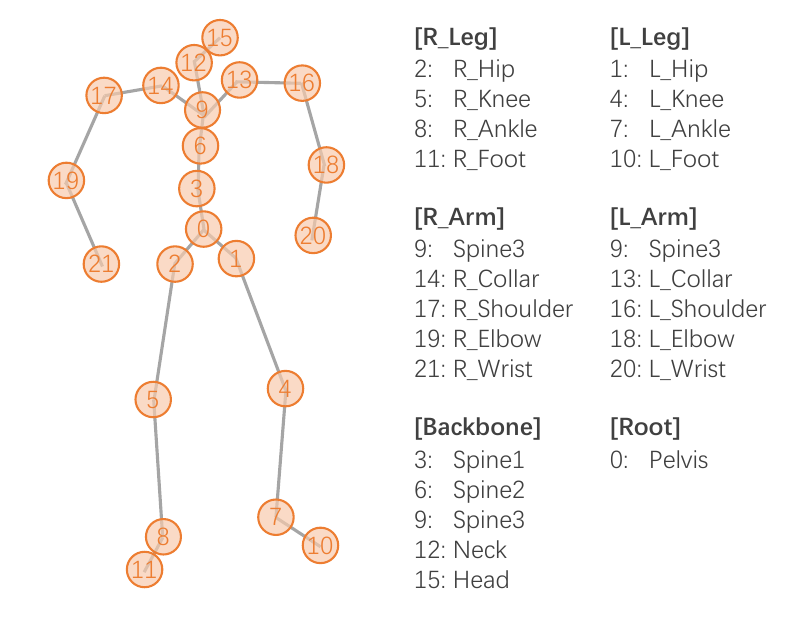}
        \caption{
        ParCo's 6-Part Division for SMPL Human Model.
        }
        \label{fig:suppl_smpl_division}
    \end{minipage}
    \hfill
    \begin{minipage}{0.48\textwidth}
        \centering
        \includegraphics[width=\linewidth]{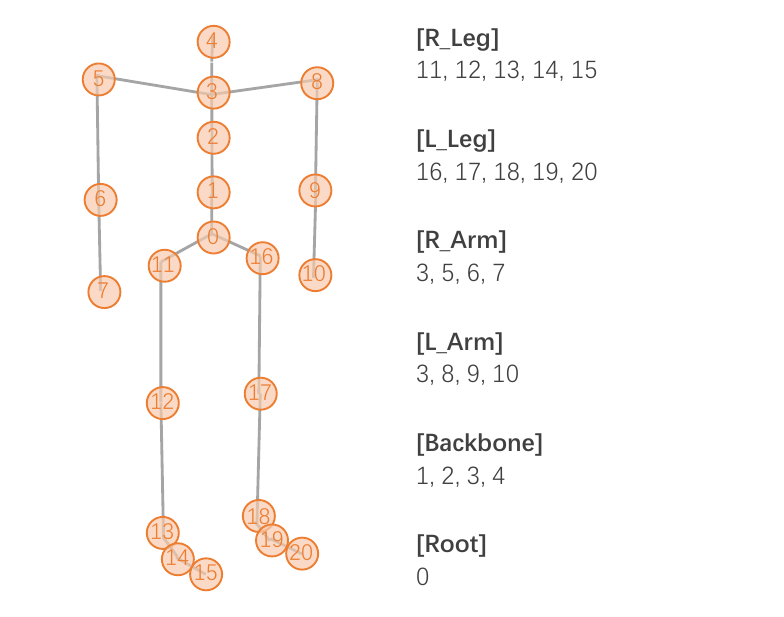}
        \caption{
        ParCo's 6-Part Division for MMM Human Model.
        }
        \label{fig:suppl_mmm_division}
    \end{minipage}
    
\end{figure}

Our ParCo divides the whole body into six parts: R.Leg, L.Leg, R.Arm, L.Arm, Backbone, and Root. Specific partitioning details for HumanML3D and KIT-ML are illustrated in Fig.~\ref{fig:suppl_smpl_division} and Fig.~\ref{fig:suppl_mmm_division}.
Both R.Arm and L.Arm include the 9-th joint for HumanML3D (3-th joint for KIT-ML).
The inclusion of the joint in both arms is due to its role as a key point connecting the arms and the backbone, providing positional information for the arms relative to this connection point. 
When reconstructing the whole-body motion from part motions, we obtain three predictions of this joint from R.Arm, L.Arm, and Backbone.
We use the average of these three values as the final prediction.

\subsubsection{Upper and Lower Body Division}
The upper-and-lower-body division is proposed by SCA~\cite{ghosh2021synthesis}, which divides the human body into upper and lower halves, both containing the backbone joints. 
In our ablation experiments, we perform the upper-and-lower-body division on the HumanML3D as,
\begin{itemize}
\setlength{\itemsep}{0.2em}
    \item Upper: 9, 14, 17, 19, 21, 13, 16, 18, 20, 0, 3, 6, 12, 15
    \item Lower: 0, 2, 5, 8, 11, 1, 4, 7, 10, 3, 6, 9, 12, 15
\end{itemize}
where the numbers denote the joint number.
It is noteworthy that, despite SCA dividing whole-body into upper and lower body for motion generation, its generations of upper body motion and lower body motion are entirely independent and lack coordination compared to our method.



\section{Details of Text-Length-Based Splits}
\label{sec:splits_details}
In order to investigate the synthetic performance given textual descriptions of different lengths, we divide the HumanML3D test set into four splits based on the length of textual descriptions. 
The test set contains a total of 4,384 motions, each motion is described by multiple textual descriptions. Following ~\cite{zhang2023t2m} and ~\cite{guo2022generating}, we set the maximum motion length to 196 and the minimum length to 40, resulting in a total of 12,635 motion-text pairs.
The distribution of these pairs, sorted by text length, is shown in Fig.~\ref{fig:suppl_text_length_distribution}.
We further divide these pairs into four subsets (0-25\%, 25-50\%, 50-75\%, 75-100\%) from short to long.
The details are shown in Table.~\ref{tab:humanml3d_4splits_analysis}, including the shortest/longest text lengths, the average length, the number of pairs, and the percentage.

\begin{table*}[t]
\caption{
VQ-VAE Reconstruction Performance on HumanML3D and KIT-ML test sets.
}
\label{tab:vqvae_humanml3d_kit}
\centering
\resizebox{1.\linewidth}{!}
{
\begin{tabular}{ll ccccccc}
\toprule[1.25pt]
\multirow{2}{*}{{Datasets}} & \multirow{2}{*}{{Methods}} &\multicolumn{3}{c}{{R-Precision $\uparrow$}} & \multirow{2}{*}{{FID $\downarrow$}} & \multirow{2}{*}{{MM-Dist $\downarrow$}} & \multirow{2}{*}{{Diversity $\rightarrow$}} \\
\cmidrule(rl){3-5}
  & & Top-1 & Top-2 & Top-3 \\ 
    \midrule
    \multirow{4}{*}{{HumanML3D}} &
    Real Motion & $0.511^{\pm.003}$ & $0.703^{\pm.003}$ & $0.797^{\pm.002}$ & $0.002^{\pm.000}$ & $2.974^{\pm.008}$ & $9.503^{\pm.065}$ \\ 
    
    \cmidrule(lr){2-8}
    & T2M-GPT & ${0.501^{\pm.002}}$ & ${0.692^{\pm.002}}$ & ${0.785^{\pm.002}}$ & ${0.070^{\pm.001}}$ & ${{3.072^{\pm.009}}}$ & ${9.593^{\pm.079}}$ \\

    & Up\&Low & ${0.488^{\pm.002}}$ & ${0.683^{\pm.002}}$ & ${0.780^{\pm.002}}$ & ${0.066^{\pm.001}}$ & ${{3.100^{\pm.007}}}$ & $\bm{9.581^{\pm.062}}$ \\
    
    \rowcolor{aliceblue!60} & ParCo (Ours) & $\bm{0.503^{\pm.003}}$ & $\bm{0.693^{\pm.003}}$ & $\bm{0.790^{\pm.002}}$ & $\bm{0.021^{\pm.000}}$ & $\bm{3.019^{\pm.007}}$ & ${9.411^{\pm.086}}$ \\

    \midrule
    \multirow{3}{*}{{KIT-ML}} &
    Real Motion & $0.424^{\pm.005}$ & $0.649^{\pm.006}$ & $0.779^{\pm.006}$ & $0.031^{\pm.004}$ & $2.788^{\pm.012}$ & $11.08^{\pm.097}$ \\ 

    \cmidrule(lr){2-8}
    & T2M-GPT & $0.399^{\pm.005}$ & $0.614^{\pm.005}$ & $0.740^{\pm.006}$ & $0.472^{\pm.011}$ & $2.986^{\pm.027}$ & $\bm{10.994^{\pm.120}}$ \\

    \rowcolor{aliceblue!60} & ParCo (Ours) & $\bm{0.407^{\pm.007}}$ & $\bm{0.629^{\pm.005}}$ & $\bm{0.760^{\pm.004}}$ & $\bm{0.311^{\pm.006}}$ & ${\bm{2.892^{\pm.016}}}$ & ${10.987^{\pm.081}}$ \\
    
\bottomrule[1.25pt]
\end{tabular}
}
\end{table*}

\newfloat{figtab}{htb}{fgtb}
\makeatletter
  \newcommand\figcaption{\def\@captype{figure}\caption}
  \newcommand\tabcaption{\def\@captype{table}\caption}
\makeatother

\begin{minipage}{0.45\textwidth}
    \tabcaption{
    Statistics of Text-Length-Based Splits.
    }
    \label{tab:humanml3d_4splits_analysis}
    \centering
    \resizebox{1.\linewidth}{!}
    {
    \begin{tabular}{l cccc}
    \toprule[1.25pt]
    Statistics & $0-25\%$ & $25-50\%$ & $50-75\%$ & $75-100\%$ \\
        \midrule
        Min length     & $4$    & $8$    & $11$   & $16$   \\
        Max length     & $7$    & $10$   & $15$   & $72$   \\
        Avg length     & $6.0$  & $8.9$  & $12.8$ & $22.3$ \\
        Total Count    & $3210$ & $2936$ & $3096$ & $3294$ \\
        Percentage(\%) & $25.6$ & $23.4$ & $24.7$ & $26.3$ \\
    \bottomrule[1.25pt]
    \end{tabular}
    }
\end{minipage}
\hfill
\begin{minipage}{0.45\textwidth}
       \centering
       \includegraphics[width=\linewidth]{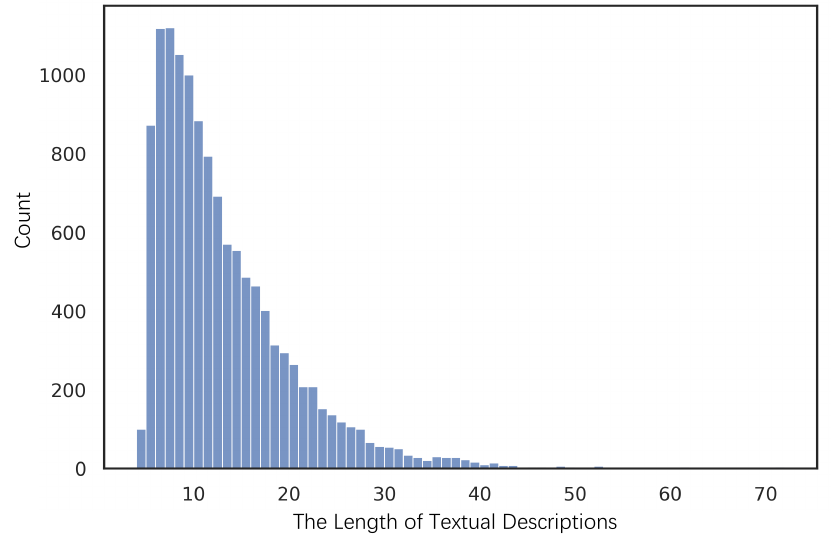}
       \figcaption{
       Distribution of Counts for Text Length.
       }
       \label{fig:suppl_text_length_distribution}
\end{minipage}

\section{VQ-VAE Reconstruction Performance}
\label{sec:vqvae_performance}
The reconstruction performance of VQ-VAEs is presented in Table.~\ref{tab:vqvae_humanml3d_kit}.
Specifically, we integrate the reconstructions of part motions into the whole-body motion for evaluation.
We conduct the ablation study of reconstruction performance with different partitioning methods on HumanML3D. 
The results indicate that the performance of our ParCo's 6 small VQ-VAE for part motion reconstruction surpasses the upper-and-lower-body division, and outperforms the baseline~\cite{zhang2023t2m} which employs a large-parameter VQ-VAE for whole-body motion.

\section{Additional Training Details}
\label{sec:additional_training_details}
During training VQ-VAE, we employ the velocity reconstruction auxiliary loss to assist training, following T2M-GPT~\cite{zhang2023t2m} and SCA~\cite{ghosh2021synthesis}. 
We use the last training checkpoint of VQ-VAE for the subsequent transformer's training. 
For the transformer, we select the checkpoint with the lowest FID during training for text-to-motion evaluation. 
Additionally, we use the decoder of transformer~\cite{vaswani2017attention} as our text-to-motion generator.
The decoder of transformer achieves autoregressive prediction by masking the upper triangle of the self-attention map.
To enhance the robustness of synthesis, we utilize the Corrupted Sequence~\cite{zhang2023t2m} strategy to augment motion sequences.
Inspired by MAE~\cite{he2022masked}, we also introduce masked part modeling, a conceptually simple yet effective approach,
to enhance part relation learning for coordinated motion generation. Specifically, we randomly replace a portion of body parts at each moment with mask tokens and force the remaining parts to predict them.
Our ParCo is trained on a single A100 GPU for a total duration of 72.8 hours (20.5 hours for stage 1 and 52.3 hours for stage 2).


\begin{figure*}[tbp]
   \centering
   \includegraphics[width=\linewidth]{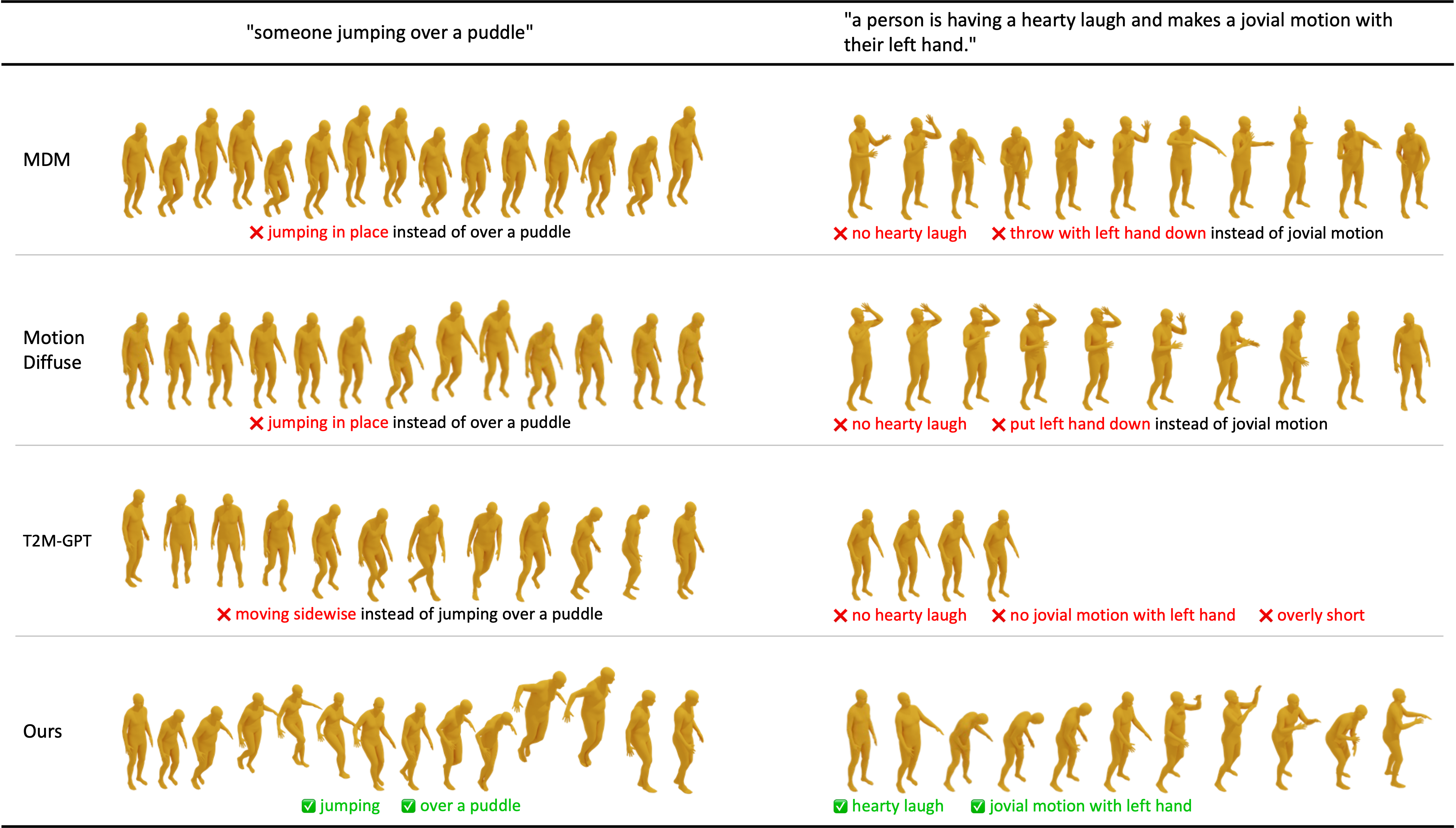}
   \caption{
   \textbf{Additional qualitative comparison with existing methods.}
    \textcolor[RGB]{0,196,12}{Green} indicates the motion is consistent with the text description.
    \textcolor{red}{Red} indicates the text description lacks the corresponding motion or got the wrong motion.
   }
   \label{fig:suppl_additional_qualitative_results}
\end{figure*}

\section{Additional Qualitative Results}
\label{sec:additional_qualitative_results}

Additional qualitative results are presented in Fig.~\ref{fig:suppl_additional_qualitative_results}. The motions are generated according to text prompts from HumanML3D test set. 
These results demonstrate that our method can generate realistic and coordinated motions aligned with the text.